\documentclass[journal,transmag]{IEEEtran}
\usepackage{times}
\usepackage{graphicx} 
\usepackage{subfigure}
\usepackage{enumerate}
\usepackage{amssymb}
\usepackage{amsmath}
\usepackage{algorithm}
\usepackage{algorithmic}
\usepackage{multirow}
\usepackage{url}
\usepackage{bm}
\usepackage[framed,numbered]{mcode}
\newtheorem{theo}{Theorem}
\newtheorem{lem}{Lemma}
\DeclareMathOperator*{\Tr}{Tr}

\begin{document}
\title{Smoothed Low Rank and Sparse Matrix Recovery by \\ Iteratively Reweighted Least Squares Minimization}
\author{Canyi Lu,~
        Zhouchen Lin,~\IEEEmembership{Senior Member,~IEEE},
        and~Shuicheng Yan,~\IEEEmembership{Senior Member,~IEEE}
\thanks{Copyright (c) 2014 IEEE. Personal use of this material is permitted. However, permission to use this material for any other purposes must be obtained from the IEEE by sending a request to pubs-permissions@ieee.org.
	
	This research is supported by the Singapore National Research Foundation under its International Research Centre @Singapore Funding Initiative and administered by the IDM Programme Office. Z. Lin is supported by NSF China (grant nos. 61272341 and 61231002), 973 Program of China (grant no. 2015CB3525) and MSRA Collaborative Research Program.
	
	C. Lu and S. Yan are with the Department of Electrical and Computer Engineering, National University of Singapore, Singapore (e-mails: canyilu@gmail.com; eleyans@nus.edu.sg).}
\thanks{Z. Lin is with the Key Laboratory of Machine Perception (MOE), School of EECS, Peking University, China (e-mail: zlin@pku.edu.cn).}}


\maketitle
\begin{abstract}
This work presents a general framework for solving the low rank and/or sparse matrix minimization problems, which may involve multiple non-smooth terms. The Iteratively Reweighted Least Squares (IRLS) method is a fast solver, which smooths the objective function and minimizes it by alternately updating the variables and their weights. However, the traditional IRLS can only solve a sparse only or low rank only minimization problem with squared loss or an affine constraint. This work generalizes IRLS to solve joint/mixed low rank and sparse minimization problems, which are essential formulations for many tasks. As a concrete example, we solve the Schatten-$p$ norm and $\ell_{2,q}$-norm regularized Low-Rank Representation (LRR) problem by IRLS, and theoretically prove that the derived solution is a stationary point (globally optimal if $p,q\geq1$). Our convergence proof of IRLS is more general than previous one which depends on the special properties of the Schatten-$p$ norm and $\ell_{2,q}$-norm. Extensive experiments on both synthetic and real data sets demonstrate that our IRLS is much more efficient.
\end{abstract}
\IEEEpeerreviewmaketitle

\IEEEpeerreviewmaketitle

%

\section{Introduction}
\label{sec:intro}
\IEEEPARstart{I}n recent years, the low rank and sparse matrix learning problems have been hot research topics and lead to broad applications in computer vision and machine learning, such as face recognition \cite{SRC}, collaborative filtering \cite{weimer2007cofirank}, background modeling \cite{RPCA}, and subspace segmentation \cite{lu2013correlation,LRR}. The $\ell_1$-norm and nuclear norm are popular choices for sparse and low rank matrix minimizations with theoretical guarantees and competitive performance in practice. The models can be formulated as a joint low rank and sparse matrix minimization problem as follow:
\begin{equation}
\label{Eq_general}
\min_\textbf{x} \sum_{i=1}^T \mathcal{F}_i(\mathcal{A}_i(\textbf{x})+\textbf{b}_i),
\end{equation}
where $\textbf{x}$ and $\textbf{b}_i$ can be either vectors or matrices, $\mathcal{F}_i$ is a convex function  (the Frobenius norm $||M||_F^2=\sum_{ij}M_{ij}^2$; nuclear norm $||M||_*=\sum_{i}\sigma_i(M)$, the sum of all singular values of a matrix; $\ell_1$-norm $||M||_1=\sum_{ij}|M_{ij}|$; and $\ell_{2,1}$-norm $||M||_{2,1}=\sum_j||M_j||_2$, the sum of the $\ell_2$-norm of each column of a matrix) and $\mathcal{A}_i: \mathbb{R}^{d}\rightarrow\mathbb{R}^m$ is a linear mapping. In this work, we further consider the nonconvex Schatten-$p$ norm $||M||_{S_p}^p=\sum_i\sigma^p(M)$, $\ell_p$-norm $||M||_p^p=\sum_{ij}|M_{ij}|^p$ and $\ell_{2,p}$-norm $||M||_{2,p}^p=\sum_j||M_j||_2^p$ with $0<p<1$ for pursuing lower rank or sparser solutions.

Problem (\ref{Eq_general}) is general which involves a wide range of problems, such as Lasso \cite{lasso}, group Lasso \cite{yuan2006model}, trace Lasso \cite{lu2013correlation}, matrix completion \cite{candes2010matrix}, Robust Principle Component Analysis (RPCA) \cite{RPCA} and Low-Rank Representation (LRR) \cite{LRR}. In this work, we aim to propose a general solver for (\ref{Eq_general}). For the ease of discussion, we focus on the following two representative problems,
\begin{equation}
\label{Eq_RPCA}
    \text{RPCA:}\ \ \ \ \ \ \ \  \min_{Z,E}  \  ||Z||_* + \lambda ||E||_1, \ \ \text{s.t.} \   X=Z+E,
\end{equation}
\begin{equation}\label{Eq_LRR}
     \text{LRR:} \ \ \ \ \ \min_{Z,E}  \   ||Z||_* + \lambda ||E||_{2,1},  \ \ \text{s.t.} \  X=XZ+E,
\end{equation}
where $X\in\mathbb{R}^{d\times n}$ is a given data matrix, $Z$ and $E$ are with compatible dimensions and $\lambda>0$ is the model parameter. Notice that these problems can be reformulated as unconstrained problems (by representing $E$ by $Z$) as that in problem ($\ref{Eq_general}$).

\subsection{Related Works}
The sparse and low rank minimization problems can be solved by various methods, such as Semi-Definite Programming (SDP) \cite{jaggi2010simple}, Accelerated Proximal Gradient (APG) \cite{APG}, and Alternating Direction Method (ADM) \cite{ALMlin}. However, SDP has a complexity of $O(n^6)$ for an $n\times n$ sized matrix, which is unbearable for large scale applications. APG requires that at least one term of the objective function has Lipschitz continuous gradient. Such an assumption is violated in many problems, e.g., problem (\ref{Eq_RPCA}) and (\ref{Eq_LRR}). Compared with SDP and APG, ADM is
the most widely used one. But it usually requires introducing several auxiliary variables corresponding to non-smooth terms. The auxiliary variables may slow down the convergence, or even lead to divergence when there are too many variables. Linearized ADM (LADM) \cite{LADMAP} may reduce the number of auxiliary variables, but suffer the same convergence issue. The work \cite{LADMAP} proposes an accelerated LADM with Adaptive Penalty (LADMAP) with lower per-iteration cost. However, the accelerating trick is special for the LRR problem. And thus are not general for other problems. Another drawback for many low rank minimization solvers is that they have to perform the soft singular value thresholding:
\begin{equation}\label{Eq_svt}
\min_Z \lambda||Z||_*+\frac{1}{2}||Z-Y||_F^2,
\end{equation}
as a subproblem. Solving (\ref{Eq_svt}) requires computing the partial SVD of $Y$. If the rank of the solution is not sufficiently low, computing the partial SVD of $Y$ is not faster than computing the full SVD of $Y$ \cite{ALMlin}.


In this work, we aim to solve the general problem (\ref{Eq_general}) without introducing auxiliary variables and also without computing SVD. The key idea is to smooth the objective function by introducing regularization terms. Then we propose the Iteratively Reweighted Least Squares (IRLS) method for solving the relaxed smooth problem by alternately updating a variable and its weight. Actually, the reweighting methods have been studied for the $\ell_p$ ($0<p\leq1$) minimization problem \cite{lu2014proximal, chartrand2008iteratively, candes2008enhancing}. Several variants have been proposed with much theoretical analysis \cite{foucart2009sparsest,zhaoreweighted}. Usually, IRLS converges exponentially fast (linear convergence) \cite{IRLSconvergencerate}, and numerical results have indicated that it leads to a sparse solution with better recovery performance. The reweighting method has also been applied for low rank minimization recently \cite{lu2014generalized,lu2015generalized,IRLSrank}. However, the problems that can be solved by iteratively reweighted algorithm are still very limited. Previous works are only able to minimize the single $\ell_1$-norm only or nuclear norm only with squared loss or an affine constraint. Thus they cannot solve (\ref{Eq_general}) whose objective function contains two or more non-smooth terms, such as robust matrix completion \cite{hsu2011robust} and RPCA \cite{RPCA}. Also, previous convergence proofs, based on the special properties of $\ell_p$-norm and Schatten-$p$ norm, are not general, and thus limit the application of IRLS. Actually, many other different nonconvex surrogate functions of $\ell_0$-norm have been proposed, e.g. the logarithm fcuntion \cite{candes2008enhancing}. We will generalize IRLS for solving problem (\ref{Eq_general}) with more general objective functions.

\subsection{Contributions}
In summary, the contributions of this paper are as follows.
\begin{itemize}
\item For solving problem (\ref{Eq_general}) with the objective function as the low rank and sparse matrix minimization, we first introduce regularization terms to smooth the objective function, and solve the relaxed problem by the Iteratively Reweighted Least Squares (IRLS) method. This is actually one of the future works mentioned in \cite{IRLSrank}.

\item We take the Schatten-$p$ norm and $\ell_{2,q}$-norm regularized LRR problem as a concrete example to introduce the IRLS algorithm and theoretically prove that the obtained solution by IRLS is a stationary point. It is globally optimal when $p,q\geq1$. Based on our general proof, we further show some other problems which can also be solved by IRLS.
\item Numerical experiments demonstrate the effectiveness of the proposed IRLS algorithm by comparing with the state-of-the-art ADM family algorithms. IRLS is much more efficient since it avoids SVD completely.
\end{itemize}

\section{Smoothed Low Rank Representation}

In this section, to illustrate the smoothed low rank and sparse matrix recovery by Iteratively Reweighted Least Squares (IRLS), we take the LRR problem as a concrete example. The reason of choosing this model as an application is twofold. First, LRR is a low rank and (column) sparse minimization problem, so solving LRR is more difficult than solving RPCA by the ADM family algorithms. It is easy to extend IRLS for other low rank plus sparse matrix recovery problems based on this example. Second, LRR has become an important model with various applications in machine learning and computer vision. A fast solver is important for real applications.

%


The LRR problem (\ref{Eq_LRR}) can be reformulated as follows without the auxiliary variable $E$:
\begin{equation}
\label{Eq_LRR2}
\min_{Z\in\mathbb{R}^{n\times n}} \mathcal{J}(Z)=||Z||_{S_p}^p+\lambda ||XZ-X||_{2,q}^q,
\end{equation}
where $||M||_{S_p}^p=\sum_i\sigma_i^p(M)$ denotes the Schatten-$p$ norm of $M$, $||M||_{2,q}^q=\sum_j||M_j||_2^q$ denotes the $\ell_{2,q}$-norm of $M$. Our solver can handle the case $0<p, q<2$. Problem (\ref{Eq_LRR}) is a special case of (\ref{Eq_LRR2}) when $p=q=1$. The major challenge for solving (\ref{Eq_LRR2}) is that both two terms of the objective function are non-smooth. A simple way is to smooth both two terms by introducing regularization terms\footnote{One may use two independent regularization parameters $\mu_1$ and $\mu_2$ for Schatten-$p$ norm and $\ell_{2,q}$-norm, respectively.}:
\begin{equation}
\label{Eq_SmoothLRR1}
\min_Z \mathcal{J}(Z,\mu)=\left \| \begin{bmatrix} Z \\
\mu I\end{bmatrix} \right \|_{S_p}^p+\lambda \left \| \begin{bmatrix} XZ-X \\
\mu\bm{1}^T \end{bmatrix} \right \|_{2,q}^q,
\end{equation}
where $\mu>0$, $I\in\mathbb{R}^{n\times n}$ is the identity matrix and $\bm{1}\in\mathbb{R}^{n}$ is the all ones vector. The terms $\mu I$ and $\mu\bm{1}^T$ make the objective function smooth (see (\ref{Eq_SmoothLRR2})). The above model is called Smoothed LRR in this work. Solving the Smoothed LRR problem instead of LRR brings several advantages.

First, $\mathcal{J}(Z,\mu)$ is smooth when $\mu>0$. This is the major difference between LRR and Smoothed LRR. Usually, a smooth objective function makes the optimization problem easier to solve.

Second, if $p, q\geq1$, $\mathcal{J}(Z)$ is convex, and so is $\mathcal{J}(Z,\mu)$. This guarantees a globally optimal solution to (\ref{Eq_SmoothLRR1}).

\begin{theo}
\label{Thm_Convex}
If $p, q\geq1$, $\mathcal{J}(Z,\mu)$ is convex w.r.t $Z$ and $\mu$. Also, for a given $\mu$, $\mathcal{J}(Z,\mu)$ is convex w.r.t $Z$.
\end{theo}

The above theorem can be easily proved by using the convexity of Schatten-$p$ norm and $\ell_{2,q}$-norm when $p, q\geq1$.

Third, $\mathcal{J}(Z,\mu)\geq\mathcal{J}(Z)$, where the equality holds if and only if $\mu=0$. Indeed,
\begin{equation*}
\begin{split}
\left \| \begin{bmatrix} Z \\
\mu I\end{bmatrix} \right \|_{S_p}^p&=\sum_{i=1}^n\left(\lambda_i(Z^TZ+\mu^2I)\right)^{\frac{p}{2}}\\
&=\sum_{i=1}^n\left(\lambda_i(Z^TZ)+\mu^2\right)^{\frac{p}{2}} \\
&\geq \sum_{i=1}^n\left(\lambda_i(Z^TZ)\right)^{\frac{p}{2}}=||Z||_{S_p}^p,
\end{split}
\end{equation*}
where $\lambda_i(M)$ denotes the $i$-th (ordered) eigenvalue of a matrix $M$. That is to say, $\mathcal{J}(Z)$ is majorized by
$\mathcal{J}(Z,\mu)$ with a given $\mu$. Decreasing $\mathcal{J}(Z,\mu)$ tends to decrease $\mathcal{J}(Z)$.

Furthermore, for any given $\epsilon>0$, there exists $\mu>0$, such that $\mathcal{J}(Z,\mu)\leq\mathcal{J}(Z)+\epsilon$. Suppose $Z^*_o$ and $Z^*$ are the optimal solutions to (\ref{Eq_LRR2}) and (\ref{Eq_SmoothLRR1}), respectively. Then we have
\begin{equation*}
0\leq\mathcal{J}(Z^*)-\mathcal{J}(Z^*_o)\leq\mathcal{J}(Z^*,\mu)-\mathcal{J}(Z^*_o,\mu)+\epsilon\leq\epsilon.
\end{equation*}
We say that the solution $Z^*$ to (\ref{Eq_SmoothLRR1}) is $\epsilon$-optimal to (\ref{Eq_LRR2}).

\begin{algorithm}[t]
\caption{Solving Smoothed LRR Problem (\ref{Eq_SmoothLRR1}) by IRLS}
\textbf{Input:} Data matrix $X\in\mathbb{R}^{m\times n}$, $\lambda>0$, $\epsilon>0$. \\
\textbf{Initialize:} $t=0$, $M_t=N_t=I\in \mathbb{R}^{n\times n}$, and $\mu>0$. \\
\textbf{while} not converged \textbf{do}
\begin{enumerate}
  \item Update $Z_{t+1}$ by solving the following problem
  	\begin{equation}
  	\label{Eq_upZ}
		pZM_t+\lambda qX^T(XZ-X)N_t=0.
	\end{equation}
  \item Update the weight matrices $M_{t+1}$ and $N_{t+1}$ separately by
  	\begin{equation}
  	\label{Eq_upMM} 	
		M_{t+1}=(Z_{t+1}^TZ_{t+1}+\mu^2I)^{\frac{p}{2}-1},
	\end{equation}
	\begin{equation}
	\label{Eq_upNN}
		(N_{t+1})_{ij}=
			\begin{cases}
			(||(XZ_{t+1}-X)_i||_2^2+\mu^2)^{\frac{q}{2}-1}, & i=j, \\
			0, \quad	 & i\neq j.
			\end{cases}
	\end{equation}
  \item $t=t+1$.
  \item If $||Z_{t+1}-Z_t||_{\infty}\leq\epsilon$, break.
\end{enumerate}
\textbf{end while}
\label{Alg_IRLS}
\end{algorithm}

\section{IRLS Algorithm}
In this section, we show how to solve (\ref{Eq_SmoothLRR1}) by IRLS. By the fact that $||Z||_{S_p}^p=\text{Tr}((Z^TZ)^{\frac{p}{2}})$, (\ref{Eq_SmoothLRR1}) can be reformulated as follows:
\begin{equation}\label{Eq_SmoothLRR2}
\min_{Z} \text{Tr}(Z^TZ+\mu^2I)^{\frac{p}{2}}+\lambda\sum_{i=1}^{n}(||(XZ-X)_i||_2^2+\mu^2)^{\frac{q}{2}},
\end{equation}
where $(M)_i$ or $M_i$ denotes the $i$-th column of matrix $M$. Let $\mathcal{L}(Z)=\text{Tr}(Z^TZ+\mu I)^{\frac{p}{2}}$ and $\mathcal{S}(Z)=\sum_{i=1}^{n}(||(XZ-X)_i||_2^2+\mu^2)^{\frac{q}{2}}$. Then $\mathcal{J}(Z,\mu)=\mathcal{L}(Z)+\lambda\mathcal{S}(Z)$.

The derivative of $\mathcal{L}(Z)$ is
\begin{equation*}
\frac{\partial\mathcal{L}}{\partial Z}=pZ(Z^TZ+ \mu^2 I)^{\frac{p}{2}-1}\triangleq pZM,
\end{equation*}
where $M=(Z^TZ+ \mu^2 I)^{\frac{p}{2}-1}$ is the weight matrix corresponding to $\mathcal{L}(Z)$. Note that $M$ can be computed without SVD \cite{higham2008functions}.


For the derivative of $\mathcal{S}(Z)$, consider the column-wise differentiation for each $i=1,\cdots,n$,
\begin{equation*}
\frac{\partial\mathcal{\mathcal{S}}}{\partial Z_i}=\frac{q(X^TXZ_i-X^TX_i)}{(||(XZ-X)_i||_2^2+\mu^2)^{1-\frac{q}{2}}}.
\end{equation*}
That is to say, $\frac{\partial\mathcal{\mathcal{S}}}{\partial Z}=qX^T(XZ-X)N$, where $N$ is the weight matrix corresponding to $\mathcal{S}(Z)$. It is a diagonal matrix with the $i$-th diagonal entry being $N_{ii}=(||(XZ-X)_i||_2^2+\mu^2)^{\frac{q}{2}-1}$.

By setting the derivative of $\mathcal{J}(Z,\mu)$ with respect to $Z$ to zero, we have
\begin{equation*}
\frac{\partial\mathcal{\mathcal{J}}}{\partial Z}=pZM+\lambda q X^T(XZ-X)N=0,
\end{equation*}
or equivalently,
\begin{equation}
\label{Eq_lyap}
\lambda qX^TXZ+pZ(MN^{-1})=\lambda qX^TX.
\end{equation}
Eqn (\ref{Eq_lyap}) is the well known Sylvester equation, which cost $O(n^3)$ for a general solver. But if $X^TX$ has certain structure, the costs may likely be $O(n^2)$ \cite{benner2009adi}. We use the Matlab command \mcode{lyap} to solve (\ref{Eq_lyap}) in this work.

Notice that both $M$ and $N$ depend only on $Z$. They can be computed if $Z$ is fixed. If the weight matrices $M$ and $N$ are fixed, $Z$ can be obtained by solving (\ref{Eq_lyap}). This fact motivates us to solve (\ref{Eq_SmoothLRR2}) by iteratively updating $Z$ and $\{M,N\}$. This optimization method is called Iteratively Reweighted Least Squares (IRLS), which is shown in Algorithm \ref{Alg_IRLS}. IRLS separately treats the weight matrices $M$ and $N$, which correspond to the low rank and sparse terms, respectively.

It is easy to see the per-iteration complexity of IRLS for the smoothed LRR problem (\ref{Eq_SmoothLRR1}) is $O(n^3)$. Such cost is the same as APG, ADM, LADM, and LADMAP. APG solves an approximated unconstraint problem of LRR. Thus its solution is not optimal to (5) or (6) \cite{LADMAP}. The traditional ADM does not guarantee to converge for LRR with three variables. Both LADM and LADMAP lead to the optimal solution of LRR. But their convergence rates are sublinear, i.e., $O(1/K$), where $K$ is the number of iterations. Usually, IRLS converges much faster than the ADM type methods and it avoids computing SVD in each iteration. Though the convergence rate of IRLS is not established, our experiments show that it tends to converge linearly. The state-of-the-art method, accelerated LADMAP \cite{LADMAP}, costs only $O(n^2r)$, where $r$ is the predicted rank of $Z$. It may be faster than our IRLS when the rank of $Z$ is sufficiently low. However, the rank of $Z$ depends on the choice of the parameter $\lambda$, which is usually tuned to achieve good performance of the application. As observed in the experiments shown later, IRLS outperforms the accelerated LADMAP on several real applications.

It is worth mentioning that though we present IRLS for LRR, it can also be used for many other problems, including the structured Lassos (e.g., group Lasso \cite{yuan2006model}, overlapping/non-overlapping group Lasso \cite{jacob2009group}, and tree structured group Lasso \cite{kim2010tree}), robust matrix completion \cite{hsu2011robust} and RPCA \cite{RPCA}. Though it is difficult to give a general IRLS algorithm for all these problems. The main idea is quite similar. The first step is to smooth the objective function like that in (\ref{Eq_SmoothLRR1}). Table \ref{tab_norms} shows the smoothed versions of some popular norms. Other related norms, e.g., overlapping group Lasso, can be smoothed in a similar way. Then we are able to compute the derivatives of the smooth functions. The derivatives can be rewritten as a simple function of the main variable $Z$ or $z$ by introducing an auxiliary variable, i.e., the weight matrix $W$ as shown in Table \ref{tab_norms}. This will make the updating of the main variable much easier. Iteratively updating the main variable $Z$ and the weight matrix $W$ leads to the IRLS algorithm which guarantees to converge. More generally, one may use other concave function, e.g., the logarithm function \cite{candes2008enhancing}, to replance the $\ell_p$-norm in Talbe \ref{tab_norms}. The induced problems can be also solved by IRLS.

\begin{table*}[!t]
\label{tab_norms}
\caption{Som popular norms, their smoothed versions and derivative ($0<p<2$).}
\label{Tab_YaleB}
\footnotesize
\centering
\begin{tabular}{|c |c|  c| c| c |}\hline
Norm & Definition & Smoothed & Derivative   & Weight matrix        	\\\hline
$\ell_p$-norm $||z||_p^p$ 		&$\sum_{i}|z_{i}|^p$ 	& 	 $\sum_{i}(z_{i}^2+\mu^2)^{\frac{p}{2}}$ 	&	$pWz$              &	$W$ is a diagonal matrix, with $W_{ii}=(z_i^2+\mu^2)^{\frac{p}{2}-1}$	\\\hline
Nuclear norm $||Z||_{S_p}^p$		& $\sum_{i}\sigma_i^p(Z)=\text{Tr}(Z^TZ)^{\frac{p}{2}}$ 	&	 $\text{Tr}(Z^TZ+\mu^2I)^{\frac{p}{2}}$	&	$pZW$          &	$W=(Z^TZ+\mu^2I)^{\frac{p}{2}-1}$	\\\hline
nonoverlapping group Lasso 	& $\sum_{i}||z_{g_i}||_2^p$,  	&	 \multirow{2}*{$\sum_{i}(||z_{g_i}||_2^2+\mu^2)^p$}	& \multirow{2}*{$pWz$}	         & $W$ is a diagonal matrix,	$W=\text{Diag}(W_1,\cdots,W_{i},\cdots)$,		\\
$||z||_{g,p}^p$ &	$g_i$ is the index of $i$-th group & &&with each $W_i$ as $(W_i)_{jj}=(||z_{g_j}||_2^2+\mu^2)^{\frac{p}{2}-1}$\\\hline
\end{tabular}
\end{table*}


\section{Algorithmic Analysis}\label{Sec_AlgAna}
Previous iteratively reweighted algorithm minimizes the sum of a non-smooth term and squared loss, while we minimize the sum of two (or more) non-smooth terms. In this section, we provide a new convergence analysis of IRLS for non-smooth optimization. Though based on Algorithm \ref{Alg_IRLS} for solving LRR problem, our proofs are general. We first show some lemmas and prove the convergence of IRLS.

Our proofs are based on a key fact that $x^p$ is concave on $(0,\infty)$ when $0<p<1$. By the definition of concave function, we have
\begin{equation}\label{eq_conca}
y^p-x^p+py^{p-1}(x-y)\geq0, \text{ for any } x,y>0.
\end{equation}
The following proofs are also applicable to other concave functions, e.g., $\log(x)$, which is an approximation of the $\ell_0$-norm of $x$.
\begin{lem}
\label{Lem_ineq1}
Assume each column of $X\in\mathbb{R}^{m\times n}$ and $Y\in\mathbb{R}^{m\times n}$ is nonzero. Let $g_i(x)$, $i=1,\cdots,n$, be concave and differentiable functions. We have
\begin{equation}\label{eq_inlem1}
\sum_{i=1}^ng_i\left(||Y_i||_2^2\right)-g_i\left(||X_i||_2^2\right)\geq\text{Tr}\left((Y^TY-X^TX)N\right),
\end{equation}
where $N\in\mathbb{R}^{n\times n}$ is a diagonal matrix, with its $i$-th diagonal element being $N_{ii}=\nabla g_i\left(||Y_i||_2^2\right)$.
\end{lem}

By letting $g_i(x)=x^{\frac{q}{2}}$, $0<q<2$, as a special case in (\ref{eq_inlem1}), we get
\begin{equation}\label{eq_inlem11}
||Y||_{2,q}^q-||X||_{2,q}^q\geq\frac{q}{2}\text{Tr}\left((Y^TY-X^TX)N\right),
\end{equation}
where $N_{ii}=(||Y_i||_2^2)^{\frac{q}{2}-1}$.

\begin{lem}
\label{Lem_ineq2}
$\text{Tr}(X^p)$ is concave on $\mathcal{S}^n_{++}$ (the set of symmetric positive definite matrices) when $0<p<1$.
\end{lem}

Assume that $h(X)$ is concave and differentiable on $\mathcal{S}^n_{++}$. For any $X, Y\in\mathcal{S}^n_{++}$, we have
\begin{equation}\label{eq_concaveh}
h(Y)-h(X)\geq\text{Tr}\left((Y-X)^T\nabla h(Y)\right).
\end{equation}
By letting $h(X)=\text{Tr}(X^{\frac{p}{2}})$ with $0<p<2$ in (\ref{eq_concaveh}), we get
\begin{equation}\label{eq_lem2r}
\begin{split}
&\left \| \begin{bmatrix} Y \\
\mu I\end{bmatrix} \right \|_{S_p}^p-\left \| \begin{bmatrix} X \\
\mu I\end{bmatrix} \right \|_{S_p}^p\\
\geq &\frac{p}{2}\Tr\left((Y^TY-X^TX)^T(Y^TY+\mu^2I)^{\frac{p}{2}-1} \right).
\end{split}
\end{equation}

Based on the above results, we have the following convergence results of the IRLS algorithm.

\begin{theo}
\label{theo_pro}
The sequence $\{Z_t\}$ generated in Algorithm \ref{Alg_IRLS} satisfies the following properties:
\renewcommand{\theenumi}{(\arabic{enumi})}
\begin{enumerate}[(1)]
\item  $\mathcal{J}(Z_t,\mu)$ is non-increasing, i.e. $\mathcal{J}(Z_{t+1},\mu)\leq\mathcal{J}(Z_t,\mu)$;
\item The sequence $\{Z_t\}$ is bounded;
\item $\lim\limits_{t\rightarrow\infty}||Z_t-Z_{t+1}||_F=0$.
\end{enumerate}
\end{theo}
\begin{theo}
\label{theo_con}
Any limit point of the sequence $\{Z_t\}$ generated by Algorithm \ref{Alg_IRLS} is a stationary point of problem (\ref{Eq_SmoothLRR1}). If $p, q\geq1$, the stationary point is globally optimal.
\end{theo}

Though for the convenience of description, we fixed $\mu>0$ in Algorithm \ref{Alg_IRLS} and the convergence analysis. In the implementation, we decrease the value of $\mu$ in each iteration, e.g., $\mu_{t+1}=\mu_t/\rho$ with $\rho>1$. The intuition is that it shall make the Smoothed LRR problem (\ref{Eq_SmoothLRR1}) close to the LRR problem (\ref{Eq_LRR2}). It is easy to check that our proofs also hold when $\mu_t\rightarrow\mu^*>0$.

It is worth mentioning that our IRLS algorithm and convergence proofs are much more general than that in \cite{IRLSconvergencerate,IRLSrank,laiimproved}, and such extensions are nontrivial. The problems in \cite{IRLSconvergencerate} and \cite{IRLSrank} are sparse or low rank minimization problems with affine constraints. The work in \cite{laiimproved} considers the unconstrained sparse or low rank minimization problems with squared loss. Our work considers an unconstrained joint low rank an sparse minimization problem. We need to update a variable and two (can be more) weight variables, while previous IRLS methods update only one variable and one weight. Note that it is usually easy to prove the convergence with two updating variables, but difficult with more than two updating variables. Also, the proofs are totally different. In \cite{IRLSconvergencerate,IRLSrank}, due to the affine constraints (i.e. $y=Ax$), the optimal solution can be written as $x^*=x_0+z$, where $x_0$ is a feasible solution and $z$ lies in the kernel of $A$. This key property is critical for their proofs but cannot be used in our proof, and we do not rely on it. The least square loss function plays an important role in the convergence proof in \cite{laiimproved} (easy to see this from equations (2.12) and (2.13) in \cite{laiimproved}). Our proof has to handle at least two non-smooth terms (and without smooth squared loss function) simultaneously. Also previous IRLS methods use a special property of $x^p$ ($0<p<1$) based on Young's inequality, while we use the concavity of $x^p$ (see (\ref{eq_inlem1}) and Lemma \ref{Lem_ineq1}, \ref{Lem_ineq2}), which involves more general functions. Thus, IRLS can be also used if $x^p$ is replaced with other concave functions, e.g., $\log(x)$.

\section{Experiments}
\label{sec_exp}

In this section, we conduct numerical experiments on both synthetic and real data to demonstrate the efficiency of the proposed IRLS algorithm\footnote{The codes can be found at https://sites.google.com/site/canyilu/.}. We use IRLS to solve LRR and Inductive Robust Principle Component (IRPCA) \cite{bao2012inductive} problems. To compare with previous convex solvers for LRR, we set $p=q=1$ in (\ref{Eq_LRR2}). We first examine the behaviour of IRLS and its sensitivity to the regularization parameter $\mu$, and then compare the performance of IRLS with state-of-the-art methods.

\subsection{Selection of Regularization Parameter $\mu$}
\label{sec_expregu}

IRLS converges fast and leads to an accurate solution when the regularization parameter $\mu$ is chosen appropriately. We decrease $\mu$ by $\mu_{t+1}=\mu_t/\rho$ with $\rho>1$. $\mu_0$ is initialized as $\mu_0=\mu_c||X||_2$, where $||X||_2$ is the spectral norm of $X$. Thus the choice of $\mu$ depends on $\mu_c$ and $\rho$. We conduct two experiments to examine the sensitivity of IRLS to $\mu_c$ and $\rho$, respectively. The first one is to fix $\rho=1.1$ and examine different values of $\mu_c$. The second one is to fix $\mu_c=0.1$ and examine different values of $\rho$. The experiments are performed on a synthetic data set.

The synthetic data is generated by the same procedure as that in \cite{LRR,LADMAP}. We generate $k=15$ independent subspaces $\{\mathcal{S}_i\}_{i=1}^k$ whose bases $\{U_i\}_{i=1}^k$ are computed by $U_{i+1}=TU_i$, $1\leq i\leq k$, where $T$ is a random rotation matrix and $U_1\in\mathbb{R}^{d\times r}$ is a random orthogonal matrix. So each subspace has a rank of $r=5$ and the data dimension is $d=200$. We sample $n_i=20$ data vectors from each subspace by $X_i=U_iQ_i$, $1\leq i\leq k$, with $Q_i$ being an $r\times n_i$ i.i.d $\mathcal{N}(0,1)$ matrix. We randomly chose $20\%$ samples to be corrupted by adding Gaussian noise with zero mean and standard deviation $0.1||x||_2$.

\begin{figure}[!t]
\centering
\includegraphics[width=0.48\textwidth]{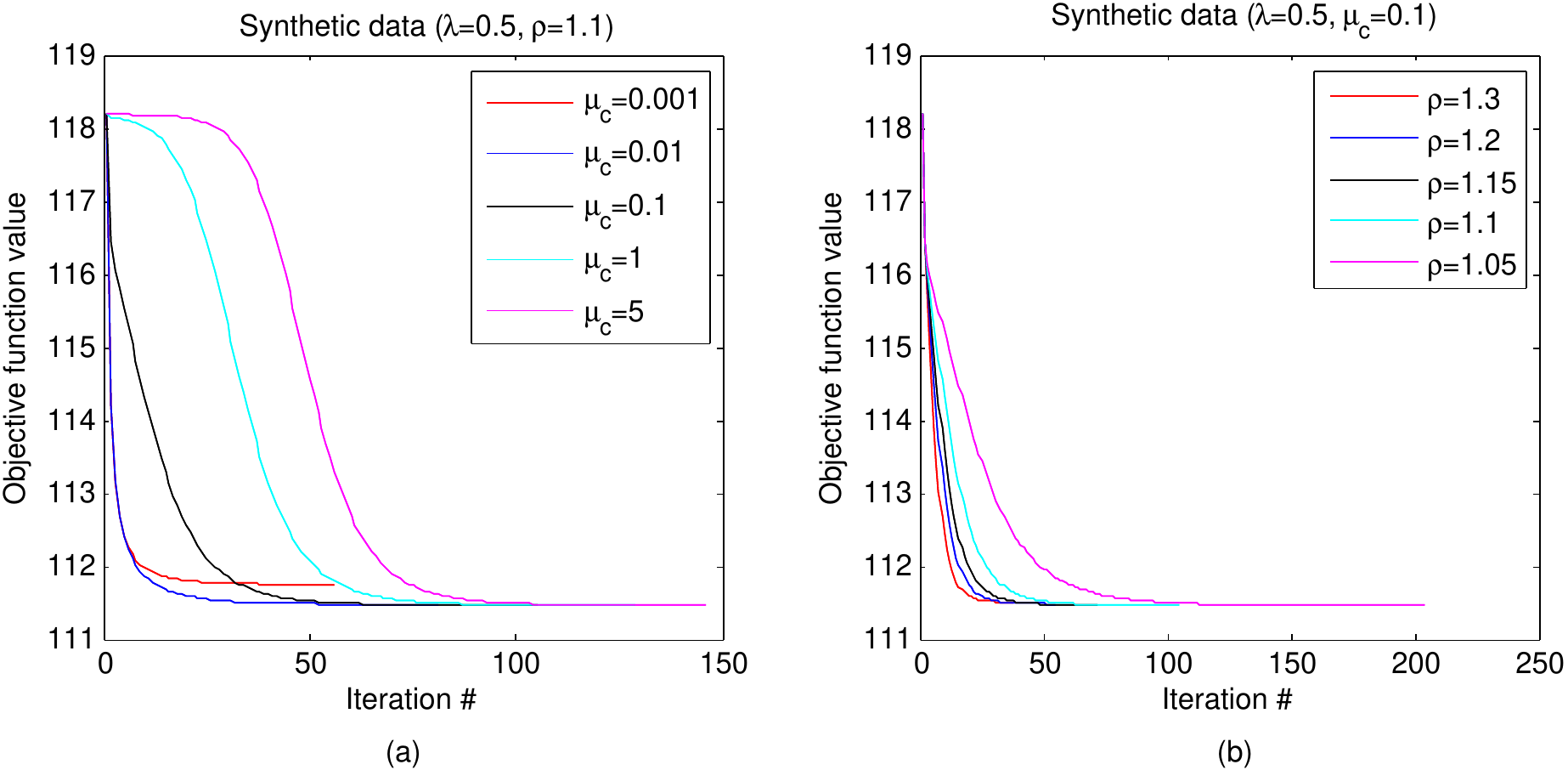}
\caption{Convergence curves of IRLS algorithm on the synthetic data with different regularization parameters $\mu_c$ and $\rho$. The LRR model parameter is $\lambda=0.5$. (a) shows the convergence curves of IRLS algorithm with different $\mu_c$ by fixing $\rho=1.1$. (b) shows the convergence curves of IRLS algorithm with different $\rho$ by fixing $\mu_c=0.1$. }
\label{fig_conv_curve_parameters}
\end{figure}

Figures \ref{fig_conv_curve_parameters} (a) and (b) show the convergence curves of IRLS with different values of $\mu_c$ and $\rho$. It is observed that a small value of $\mu_c$ will lead to an inaccurate solution in a few iterations. But a large value of $\mu_c$ will delay the convergence. Similar phenomenon can be found in the choice of $\rho$. A large value of $\rho$ will lead to fast convergence, while a small value of $\rho$ will lead to a more accurate solution. For an accurate solution, $\mu$ should not converge to 0 too fast. Thus $\mu_c$ cannot be too small and $\rho$ should not be too large. We observe that $\mu_c=0.1$ and $\rho=1.1$ work well.

\begin{figure}[!t]
\centering
\includegraphics[width=0.5\textwidth]{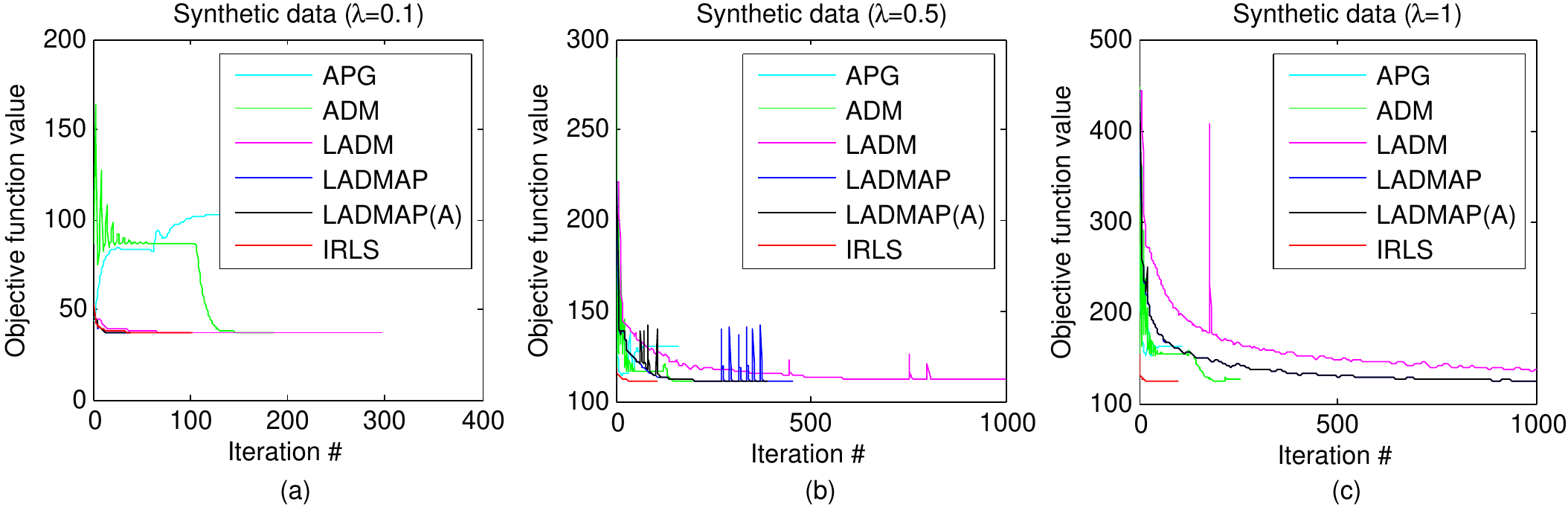}
\caption{Convergence curves of APG, ADM, LADM, LADMAP, LADMAP(A) and IRLS algorithms on the synthetic data with different LRR model parameters: (a) $\lambda=0.1$, (b) $\lambda=0.5$, and (c) $\lambda=1$.}
\label{fig_conv_curve_toydata}
\end{figure}

\subsection{LRR for Subspace Segmentation}

In this section, we present numerical results of IRLS and the other state-of-the-art algorithms, including APG, ADM, LADM \cite{LADM}, LADMAP and accelerated LADMAP \cite{LADMAP} (denoted as LADMAP(A)) to solve the LRR problem for subspace segmentation. All the ADM type methods use PROPACK \cite{larsen1998lanczos} for fast SVD computing. We implement  IRLS  algorithm by Matlab without using third party package.
For LADMAP(A), we set the maximum iteration number as 10000 (the default value is 1000). This is because LADMAP(A) is usually fast but not able to converge within 1000 iterations in some cases. Except this, we use the default parameters of all the competed methods in the released codes from Lin's homepage\footnote{http://www.cis.pku.edu.cn/faculty/vision/zlin/zlin.htm}. For IRLS, we set $\mu_0=\mu_c||X||_2=0.1||X||_2$, $\mu_{t+1}=\mu_t/\rho$ and $\rho=1.1$. All experiments are run on a PC with an Intel Core 2 Quad CPU Q9550 at 2.83GH and 8GB memory, running Windows 7 and Matlab version 8.0.

\begin{table}[!t]
\caption{Experiments on the synthetic data with different LRR model parameters. The obtained minimum, running time (in seconds) and iteration number are presented for comparison.}
\label{Tab_Toydata}
\footnotesize
\centering
\begin{tabular}{|c c c c|} \hline
\multicolumn{4}{|c|} {$\lambda=0.1$}\\
Method & Minimum & Time & Iter.  \\ \hline
APG 	 &111.481 	 &129.6 	 &312 \\
ADM 	 &37.572 	 &77.2 	 &187\\
LADM 	 &\textbf{37.571} 	 &130.3 	 &298\\
LADMAP 	 &\textbf{37.571} 	 &16.8 	 &\textbf{38}	\\
LADMAP(A) &\textbf{37.571} 	 &\textbf{2.4} 	 &\textbf{38}\\
IRLS 	 &\textbf{37.571} 	 &26.5 	 &105\\\hline\hline
\multicolumn{4}{|c|} {$\lambda=0.5$}\\
Method & Minimum & Time & Iter.  \\ \hline
APG 	 &129.022 	 &56.2 	 &160 \\
ADM 	 &\textbf{111.463} 	 &76.6 	 &199 \\
LADM 	 &111.797 	 &418.2 	 &$>$1000\\
LADMAP 	 &\textbf{111.463} 	 &175.2 	 &457\\
LADMAP(A)&\textbf{111.463} 	 &123.6 	 &391\\
IRLS 	 &\textbf{111.463} 	 &\textbf{26.4} 	 &\textbf{105}\\ \hline\hline
\multicolumn{4}{|c|} {$\lambda=1$}\\
Method & Minimum & Time & Iter.  \\ \hline
APG 	 &147.171 	 &44.0 	 &109\\
ADM 	 &124.586 	 &105.7 	 &257\\
LADM 	 &136.819 	 &578.9 	 &$>$1000\\
LADMAP 	 &124.967 	 &556.3 	 &$>$1000\\
LADMAP(A)&\textbf{123.933} 	 &1081.4 	 &1973\\
IRLS 	 &\textbf{123.933} 	 &\textbf{24.9} &	 \textbf{105}\\ \hline
\end{tabular}
\end{table}

\subsubsection{Synthetic Data Example}

We use the same synthetic data as that in Section \ref{sec_expregu}. We emphasize on the performance with different LRR model parameter $\lambda$. Usually a larger $\lambda$ leads to lower rank solution. This experiment is to test the sensitiveness of the competed methods to different ranks of the solution. Figure \ref{fig_conv_curve_parameters} shows the convergence curves corresponding to $\lambda=0.1, 0.5$ and $1$, respectively (only the results within 1000 iterations are plotted). Table \ref{Tab_Toydata} shows the detailed results, including the achieved minimum at the last iteration, the computing time and the number of iterations. It can be seen that IRLS is always faster than APG, ADM and LADM. IRLS also outperforms LADMAP and LADMAP(A) except when $\lambda=0.1$. We find that the linearized ADM methods need more iterations to converge when $\lambda$ increases. That is because when $\lambda$ is not small enough, the rank of the solution will be not small. In this case, partial SVD may not be faster than the full SVD \cite{ALMlin}. Hence using PROPACK may be unstable. Compared with LADMAP(A), IRLS is a better choice for the small-sized or high-rank problems because it completely avoids SVD.


\begin{figure}[!t]
\centering
\includegraphics[width=0.4\textwidth]{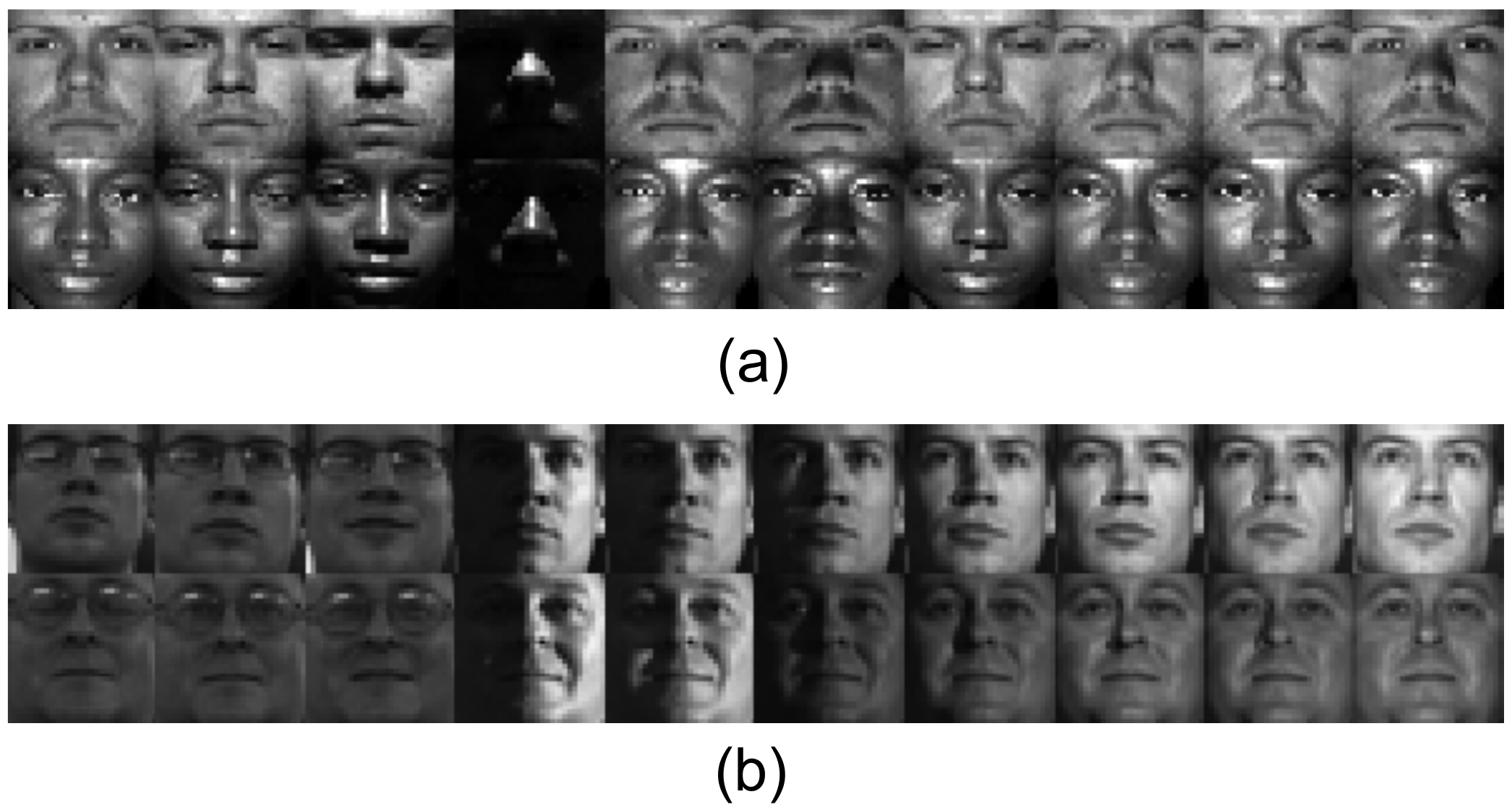}
\caption{Example face images from the (a) Yale B and (b) PIE databases.}
\label{fig_example_faces}
\end{figure}
\begin{figure}[!t]
\centering
\includegraphics[width=0.5\textwidth]{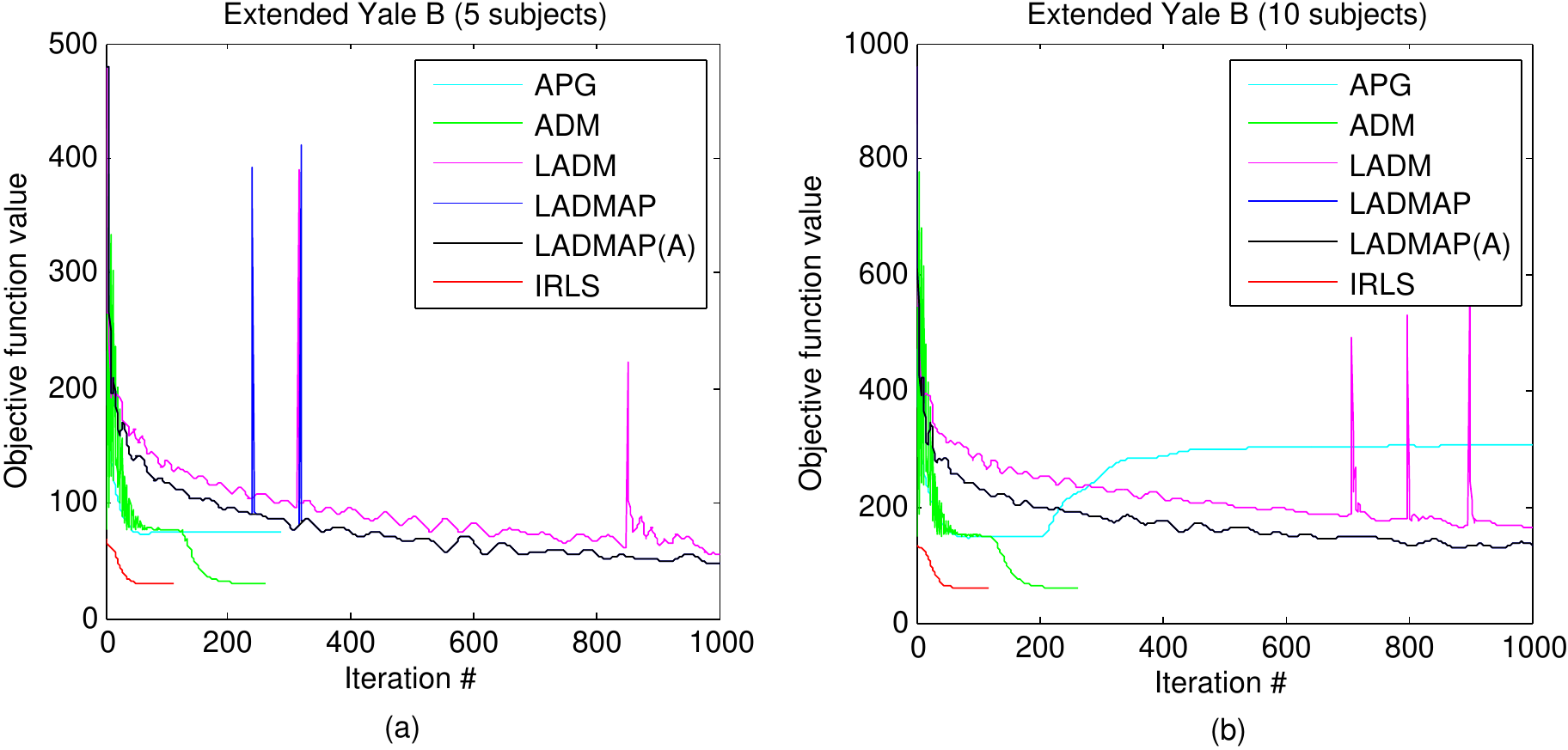}
\caption{Convergence curves of compared algorithms on two subsets of the Extended Yale B database: (a) 5 subjects and (b) 10 subjects.}
\label{fig_conv_curve_YaleB5_10}
\end{figure}
	
\subsubsection{Face Clustering}

We test the performance of all the competed methods for face clustering on the Extended Yale B database \cite{YaleBdatabase}. Some example face images are shown in Figure \ref{fig_example_faces}. There are 38 subjects in this database. We conduct two experiments by using the first 5 and 10 subjects of face images to form the data $X$ \cite{lu2012robust}. Each subject has 64 face images. These images are resized into $32\times32$ and projected onto a 30-dimensional subspace by PCA for 5 subjects clustering problem and a 60-dimensional subspace for 10 subjects clustering problem. The affinity matrix is defined as $(|Z^*|+|(Z^*)^T|)/2$, where $Z^*$ is the solution to the LRR problem obtained by different solvers. Then the Normalized Cut \cite{NormaCut} is used to produce the clustering results based on the affinity matrix. The LRR model parameter is set to $\lambda=1.5$ which leads to the best clustering accuracy.

Figure \ref{fig_conv_curve_YaleB5_10} and Table \ref{Tab_YaleB} show the performance comparison of all these methods. It can be seen that IRLS is the fastest and the most accurate method. ADM also works well but needs more iterations. The linearized methods are not efficient since they do not converge within 1000 iterations.

\subsubsection{Motion Segmentation}

We also test all the competed methods for motion segmentation on the Hopkins 155 database\footnote{http://www.vision.jhu.edu/data/hopkins155/}. This database has 156 sequences, each of which has 39 to 550 data points drawn from two or three motions. In each sequence, the data are first projected onto a 12-dimensional subspace by PCA. LRR is performed on the projected subspace, the best LRR model parameter is set to $\lambda=2.4$. Table \ref{Tab_hop} tabulates the comparison of all these methods. It can be seen that IRLS is the fastest method. LADMAP(A) is competitive with IRLS but it requires much more iterations.

\begin{table}[!t]
\caption{Comparison of face clustering by LRR by using different solvers on two subsets of the Extended Yale B database: 5 subjects and 10 subjects. The obtained minimum, running time (in seconds), number of iteration and clustering accuracy ($\%$) of each method are presented for comparison.}
\label{Tab_YaleB}
\footnotesize
\centering
\begin{tabular}{|c c  c c c |}\hline
\multicolumn{5}{|c|} {5 subjects ($\lambda=1.5$)}\\
Method 		& Minimum & Time & Iter. & Acc. \\ \hline
APG			& 74.603 	&	 117.9 	&	288              &	61.88	\\
ADM 		& 29.993 	& 	 107.5 	&	262              &	\textbf{84.69}	\\
LADM 		& 56.266 	&	 411.3 	&	$>$1000          &	\textbf{84.69}	\\
LADMAP 		& 48.178 	&	 409.0 	&	$>$1000	         &	82.81	\\
LADMAP(A) 	& 30.028 	&	 494.9 	&	8418	         &	84.14	\\
IRLS 		& \textbf{29.991} 	&	 \textbf{33.1}	&	\textbf{113}	&	\textbf{84.69}	\\ \hline\hline
\multicolumn{5}{|c|} {10 subjects ($\lambda=1.5$)}\\
Method & Minimum & Time & Iter. & Acc. \\ \hline
APG			&	305.692	&	2962.9	 	&   $>$1000		 & 	32.52   \\
ADM			&	60.001	&	705.4 	    &	262		     &  68.53	\\
LADM		&	162.488	&	2692.8 	    &	$>$1000	     &  47.34	\\
LADMAP		&	134.898	&	2681.1 	    &	$>$1000	     &  57.40	\\
LADMAP(A)	&	61.230	&	2212.3	    &	$>$10000	 &  68.44	\\
IRLS		&	\textbf{59.999}	&	\textbf{222.9}	&	\textbf{117}			 &  \textbf{69.17} \\ \hline
\end{tabular}
\end{table}
\subsection{Inductive Robust Principal Component Analysis}

Inductive Robust Principal Component Analysis (IRPCA) \cite{bao2012inductive} aims at finding a robust projection to remove the possible corruptions in data. It is done by solving the following nuclear norm regularized minimization problem
\begin{equation}\label{eq_irpca}
\min_{P} ||P||_*+\lambda||PX-X||_{1,2}.
\end{equation}
Here we use the $\ell_{1,2}$-norm $||E||_{1,2}$, sum of the $\ell_2$-norm of each row of $E$ instead of $\ell_1$-norm in \cite{bao2012inductive} to handle the data with row corruptions (caused by continuous shadow, e.g., face with glass or scarf).

\begin{figure}[!t]
\centering
\includegraphics[width=0.47\textwidth]{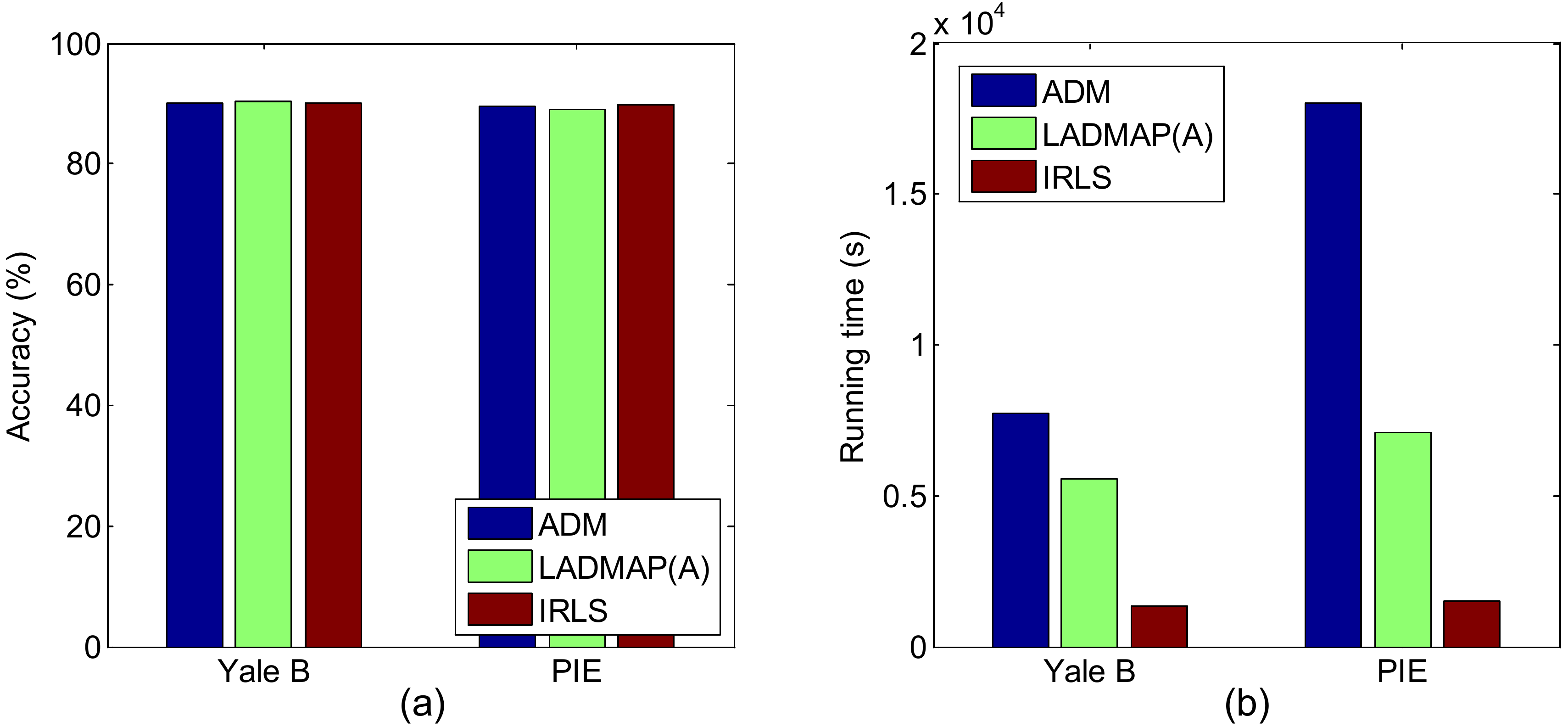}
\caption{Comparison of (a) accuracy and (b) running time of ADM, LADMAP(A) and IRLS for solving IRPCA problem on the Yale B and PIE databases.}
\label{fig_fr_accuracy_time}
\end{figure}

\begin{figure}[!t]
\centering
\includegraphics[width=0.47\textwidth]{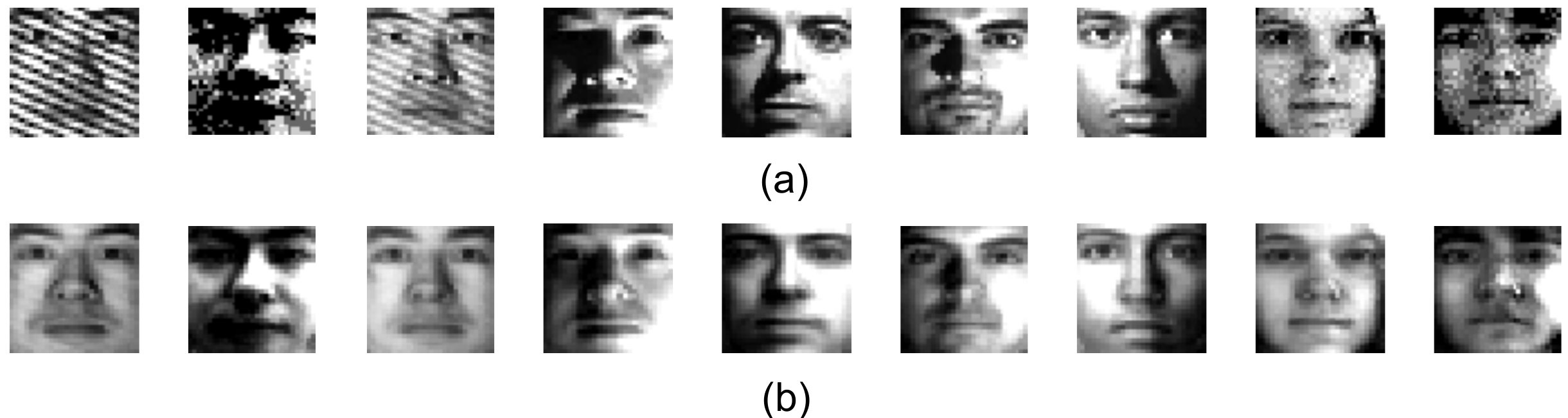}
\caption{(a) Some corrupted test face images from the Yale B database; (b) Recovered face images by IRPCA projection obtained by IRLS.}
\label{fig_face_recovery}
\end{figure}

The $\ell_{1,2}$-norm can be smoothed as $||E||_{1,2}=\sum_{i}(||E^i||_2^2+\mu^2)^{\frac{1}{2}}$, where $E^i$ denotes the $i$-th row of $E$. Thus IRLS solves (\ref{eq_irpca}) by iteratively solving
\begin{equation*}
M_tP+\lambda N_t(PX-X)X^T=0,
\end{equation*}
where $M_t=(P_tP_t^T+\mu^2I)^{-\frac{1}{2}}$ and $N_t$ is a diagonal matrix with $(N_t)_{ii}=(||(P_tX-X)^i||_2^2+\mu^2)^{-\frac{1}{2}}$. We test our IRLS by comparing with ADM in \cite{bao2012inductive} and LADMAP(A) \cite{LADMAP} for face recognition. After the projection $P$ is learned by solving (\ref{eq_irpca}) from the training data, we can use it to remove corruption from a new coming test data point. We perform experiments on two face data sets. The first one is the Extended Yale B, which consists of 38 subjects with 64 images in each subject. We randomly select 30 images for training and the rest for test. The other one is the CMU PIE face dataset \cite{sim2003cmu}, which contains more than 40,000 facial images of 68 people. The images were acquired across different poses. We use the one near frontal pose C07, which includes 1629 images. All the images are resized to $32\times 32$. For each subject, we randomly select 10 images for training, and the rest for test. The support vector machine (SVM) is used to perform classification. The recognition results are shown in Figure \ref{fig_fr_accuracy_time}. It can be seen that the recognition accuracies are almost the same by different solvers. But the running time of ADM and LAMDAP(A) is much larger than our IRLS algorithm. Figure \ref{fig_face_recovery} plots some test images recovered by IRPCA obtained by our IRLS algorithm. It can be seen that IRPCA by IRLS successfully removes the shadow and corruptions from faces.

\begin{table}[!t]
\caption{Comparison of motion segmentation by LRR by using different solvers on the Hopkins 155 database. The average running time (in seconds), average iterations number and average segmentation errors ($\%$) are reported for comparison.}
\label{Tab_hop}
\footnotesize
\centering
\begin{tabular}{|c c c c|} \hline
\multicolumn{4}{|c|} {Two Motions}      \\
Method     &    Time   &   Iter.    &  Err.     \\ \hline
APG 	   &   165.7   &   388      &  3.62     \\
ADM 	   &   100.8   &   223      &  2.48     \\
LADM 	   &   415.0   &   $>$1000  &  6.30     \\
LADMAP 	   &   368.5   &   $>$1000  &  4.50 	\\
LADMAP(A)  &   57.6    &   4668  &  \textbf{2.40}     \\
IRLS 	   &   \textbf{35.5}    &\textbf{131} &2.71 \\ \hline\hline
\multicolumn{4}{|c|} {Three Motions}    \\
Method     &    Time   &   Iter.    &  Err.     \\ \hline
APG 	   &   456.6   &   476      &  12.67    \\
ADM 	   &   222.0   &   224      &  5.45     \\
LADM 	   &   942.8   &   $>$1000  &  14.59    \\
LADMAP 	   &   883.7   &   $>$1000  &  10.12    \\
LADMAP(A)  &   89.9    &    5768    &  5.19     \\
IRLS 	   &   \textbf{84.7}    &\textbf{133}&\textbf{4.14} \\ \hline\hline
\multicolumn{4}{|c|} {All}              \\
Method     &    Time   &   Iter.    &  Err.     \\ \hline
APG 	   &   230.8   &   408      &  5.84     \\
ADM 	   &   127.9   &   223      &  3.25     \\
LADM 	   &   532.6   &   $>$1000  &  8.33     \\
LADMAP 	   &   483.3   &   $>$1000  &  5.91     \\
LADMAP(A)  &   65.7    &   4949     &  \textbf{3.19}     \\
IRLS 	   &   \textbf{46.4}    &\textbf{131} &3.20 \\ \hline
\end{tabular}
\end{table}

\section{Conclusions and Future Work}
Different from previous Iteratively Reweighted Least Squares (IRLS) algorithm which simply solved a single sparse or low rank minimization problem. We proposed a more general IRLS to solve the joint low rank and sparse matrix minimization problems. The objective function is first smoothed by introducing regularization terms. Then IRLS is applied for solving the relaxed problem. We provide a general proof to show that the solution by IRLS is a stationary point (globally optimal if the problem is convex). IRLS can also be applied to various optimization problems with the same convergence guarantee. An interesting future work is to use IRLS for solving nonconvex structured Lasso problems (e.g., $\ell_p$-norm regularized group Lasso, overlapping/non-overlapping group Lasso \cite{jacob2009group}, and tree structured group Lasso \cite{kim2010tree}).


\section*{Appendix}

\subsection{Proof of Lemma \ref{Lem_ineq1}}
\textbf{Proof.} By the definition of concave function, we have
\begin{equation*}\label{eq_prooflem1eq1}
\begin{split}
&\sum_{i=1}^ng_i\left(||Y_i||_2^2\right)-g_i\left(||X_i||_2^2\right)\\
\geq&\sum_{i=1}^n\nabla g_i\left(||Y_i||_2^2\right)\left(||Y_i||_2^2-||X_i||_2^2\right)\\
=&\text{Tr}\left((Y^TY-X^TX)N\right).
\end{split}
\end{equation*}
 $\hfill\blacksquare$

\begin{lem}
\cite{laiimproved}
\label{Lem_eng}
Given $X,Y\in\mathcal{S}^n_{++}$. Let $\lambda_1(X)\geq\lambda_2(X)\geq\cdots\geq\lambda_n(X)\geq0$ and $\lambda_1(Y)\geq\lambda_2(Y)\geq\cdots\geq\lambda_n(Y)\geq0$ be ordered eigenvalues of $X$ and $Y$, respectively. Then $\text{Tr}(XY)\geq\sum_{i=1}^n\lambda_i(X)\lambda_{n-i+1}(Y)$.
\end{lem}

\subsection{Proof of Lemma \ref{Lem_ineq2}}
\textbf{Proof.} By using Lemma \ref{Lem_eng}, for any $X, Y\in\mathcal{S}^n_{++}$, we have
\begin{equation*}
\begin{split}
\Tr(X^TY^{p-1})&\geq\sum_{i=1}^n\lambda_i(X)\lambda_{n-i+1}(Y^{p-1})\\
&=\sum_{i=1}^n\lambda_i(X)\lambda_{i}^{p-1}(Y).
\end{split}
\end{equation*}
Then we deduce
\begin{equation}\label{eq_prooflm2}
\begin{split}
&\Tr(Y^p)-\Tr(X^p)+\Tr(p(X-Y)^TY^{p-1})\\
\geq&\sum_{i=1}^n\lambda_i(Y^p)-\lambda_i(X^p)+p\lambda_i(X)\lambda_{i}^{p-1}(Y)-p\lambda_i(Y^p)\\
=&\sum_{i=1}^n\lambda_i^p(Y)-\lambda_i^p(X)+p\lambda_i^{p-1}(Y)(\lambda_i(X)-\lambda_i(Y))\\
\geq&0.
\end{split}
\end{equation}
The last inequality uses the concavity of $x^p$ with $0<p<1$ on $(0,\infty)$ in (\ref{eq_conca}). Thus $\text{Tr}(X^p)$ is concave from (\ref{eq_prooflm2}). $\hfill\blacksquare$

\subsection{Proof of Theorem \ref{theo_pro}}

\textbf{Proof.} We denote $E_t=XZ_t-X$. Since $Z_{t+1}$ solves (\ref{Eq_upZ}), we have
\begin{equation}\label{eq_proof00}
pZ_{t+1}M_t+\lambda qX^T(XZ_{t+1}-X)N_t=0.
\end{equation}
A dot product with $Z_t-Z_{t+1}$ on both side of (\ref{eq_proof00}) gives
\begin{equation*}
\begin{split}
&p(Z_t-Z_{t+1})^TZ_{t+1}M_t\\
=&-\lambda q(XZ_t-XZ_{t+1})^T(XZ_{t+1}-X)N_t\\
=&-\lambda q(E_t-E_{t+1})^TE_{t+1}N_t.
\end{split}
\end{equation*}
This together with (\ref{eq_lem2r}) gives
\begin{equation}\label{eq_proof111}
\begin{split}
&\left \| \begin{bmatrix} Z_t \\
\mu I\end{bmatrix} \right \|_{S_p}^p-\left \| \begin{bmatrix} Z_{t+1} \\
\mu I\end{bmatrix} \right \|_{S_p}^p\\
\geq&\frac{p}{2}\Tr\left(\left(Z_t^TZ_t-Z_{t+1}^TZ_{t+1}\right)^T\left(Z_t^TZ_t^T+\mu I\right)^{\frac{p}{2}-1}\right)\\
=&\frac{p}{2}\Tr\left(\left(Z_t-Z_{t+1}\right)^T\left(Z_t-Z_{t+1}\right)M_t\right)\\
&+p\Tr\left(\left(Z_t-Z_{t+1}\right)^TZ_{t+1}M_t\right)\\
=&\frac{p}{2}\Tr\left(\left(Z_t-Z_{t+1}\right)^T\left(Z_t-Z_{t+1}\right)M_t\right)\\
&-\lambda q\Tr\left((E_t-E_{t+1})^TE_{t+1}N_t\right).
\end{split}
\end{equation}
By using (\ref{eq_inlem11}), we have
\begin{equation}\label{eq_proof222}
\begin{split}
&\lambda\left \| \begin{bmatrix} E_t \\
\mu \bm{1}^T\end{bmatrix} \right \|_{2,q}-\lambda\left \| \begin{bmatrix} E_{t+1} \\
\mu \bm{1}^T\end{bmatrix} \right \|_{2,q}\\
\geq&\frac{\lambda q}{2}\Tr\left(\left(E_t^TE_t-E_{t+1}^TE_{t+1}\right)N_t\right)\\
=&\frac{\lambda q}{2}\Tr\left(\left(E_t-E_{t+1}\right)^T\left(E_t-E_{t+1}\right)N_t\right)\\
&+\lambda q\Tr\left(\left(E_t-E_{t+1}\right)^TE_{t+1}N_t\right).
\end{split}
\end{equation}
Now, combining (\ref{eq_proof111}) and (\ref{eq_proof222}) gives
\begin{equation}
\label{Eq_lempr}
\begin{split}
& \mathcal{J}(Z_t,\mu)-\mathcal{J}(Z_{t+1},\mu)\\
=&  \frac{p}{2}\text{Tr}\left((Z_t-Z_{t+1})^T(Z_t-Z_{t+1})M_t\right)\\
&+ \frac{\lambda q}{2}\text{Tr}\left((E_t-E_{t+1})^T(E_t-E_{t+1})N_t)\right)\geq0.
\end{split}
\end{equation}
The above equation implies that $\mathcal{J}(Z_t,\mu)$ is non-increasing. Then we have
\begin{equation}\label{eq_proofeng}
\begin{split}
||Z_t||_{S_p}^p&\leq\text{Tr}(Z_t^TZ_t+\mu^2)^{\frac{p}{2}}\leq\text{Tr}(M_t^{-\frac{p}{2-p}})+\lambda\text{Tr}(N_t^{-\frac{q}{2-q}})\\
&=\mathcal{J}(Z_t,\mu)\leq\mathcal{J}(Z_1,\mu)\triangleq D.
\end{split}
\end{equation}
Thus the sequence $\{Z_t\}$ is bounded. Furthermore, (\ref{eq_proofeng}) implies that the minimum eigenvalues of $M_t$ and $N_t$ satisfy
\begin{equation*}
\begin{split}
&\min\{\min_i\lambda_i(M_t),\min_i\lambda_i(N_t) \}\\
\geq& \min\{D^{\frac{p}{2-p}},\lambda^{-1} D^{\frac{q}{2-q}}\}\triangleq \theta>0.
\end{split}
\end{equation*}
By using Lemma \ref{Lem_eng}, (\ref{Eq_lempr}) implies that
\begin{equation*}
\begin{split}
 &\mathcal{J}(Z_t,\mu)-\mathcal{J}(Z_{t+1},\mu)\\
\geq&\frac{p}{2} \sum_{i=1}^n\lambda_{n-i+1}(M_t)\lambda_i\left((Z_t-Z_{t+1})^T(Z_t-Z_{t+1})\right)\\&+\frac{\lambda q}{2}\sum_{i=1}^n\lambda_{n-i+1}(N_t)\lambda_i\left((E_t-E_{t+1})^T(E_t-E_{t+1})\right)   \\
\geq & \frac{\theta}{2}\left(p||Z_t-Z_{t+1}||_F^2+\lambda q||E_t-E_{t+1}||_F^2\right).
\end{split}
\end{equation*}
Summing all the above inequalities for all $t\geq1$, we get
\begin{equation}\label{Eqn_tempppp}
\begin{split}
D=&\mathcal{J}(Z_1,\mu)\geq \frac{\theta}{2}\sum_{t=1}^\infty(p||Z_t-Z_{t+1}||_F^2+\lambda q||E_t-E_{t+1}||_F^2).
\end{split}
\end{equation}
In particular, (\ref{Eqn_tempppp}) implies that $\lim\limits_{t\rightarrow\infty}||Z_t-Z_{t+1}||_F=0$. The proof is completed. $\hfill\blacksquare$

\subsection{Proof of Theorem \ref{theo_con}}

\textbf{Proof.} If $p, q\geq1$, problem (\ref{Eq_SmoothLRR1}) is convex. The stationary point is globally optimal. Thus we only need to prove that $Z_t$ converges to a stationary point of problem (\ref{Eq_SmoothLRR1}).

The sequence $\{Z_t\}$ is bounded by Theorem \ref{theo_pro}, hence there exists a matrix $\hat{Z}$ and a subsequence $\{Z_{t_j}\}$, such that $\lim\limits_{j\rightarrow\infty} Z_{t_j}\rightarrow\hat{Z}$. Note that $Z_{t_{j}+1}$ solves (\ref{Eq_upZ}), i.e.,
\begin{equation}
\label{Eq_firstorder}
pZ_{t_{j}+1}M_{t_j}+\lambda qX^T(XZ_{t_{j}+1}-X)N_{t_j}=0.
\end{equation}
Let $j\rightarrow\infty$, (\ref{Eq_firstorder}) implies that $Z_{t_{j}+1}$ also converges to some $\tilde{Z}$. From the fact that $\lim\limits_{t\rightarrow\infty}||Z_t-Z_{t+1}||_F=0$ in Theorem \ref{theo_pro}, we have
\begin{equation*}
||\hat{Z}-\tilde{Z}||_F=\lim_{j\rightarrow\infty}||Z_{t_j}-Z_{t_{j}+1}||_F=0.
\end{equation*}
That is to say $\hat{Z}=\tilde{Z}$. Denote $\hat{Z}$ as $Z^*$, and let $j\rightarrow\infty$, (\ref{Eq_firstorder}) can be rewritten as
\begin{equation*}
pZ^*M^*+\lambda qX^T(XZ^*-X)N^*=0,
\end{equation*}
where $M^*$ and $N^*$ are defined in (\ref{Eq_upMM})(\ref{Eq_upNN}) with $Z^*$ in place of $Z_{t+1}$. Therefore, $Z^*$ satisfies the first-order optimality condition of problem (\ref{Eq_SmoothLRR1}). $\hfill\blacksquare$

{
\bibliographystyle{IEEEbib}
\bibliography{SmoothedLRR}
}
\begin{IEEEbiography}[{\includegraphics[width=1in,height=1.25in,clip,keepaspectratio]{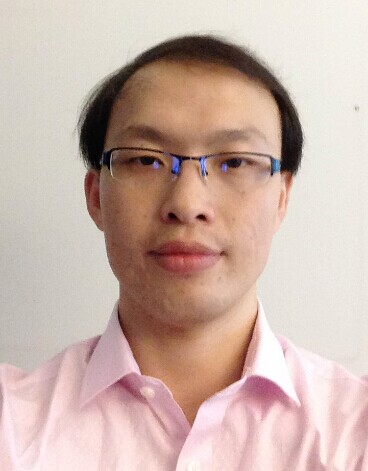}}]{Canyi Lu} received the bachelor degree in mathematics from the Fuzhou University in 2009, and the master degree in the pattern recognition and intelligent system from the University of Science and Technology of China in 2012. He is currently a Ph.D. student with the Department of Electrical and Computer Engineering at the National University of Singapore. His current research interests include computer vision, machine learning, pattern recognition and optimization. He was the winner of the Microsoft Research Asia Fellowship 2014.
\end{IEEEbiography}
\begin{IEEEbiography}[{\includegraphics[width=1in,height=1.25in,clip,keepaspectratio]{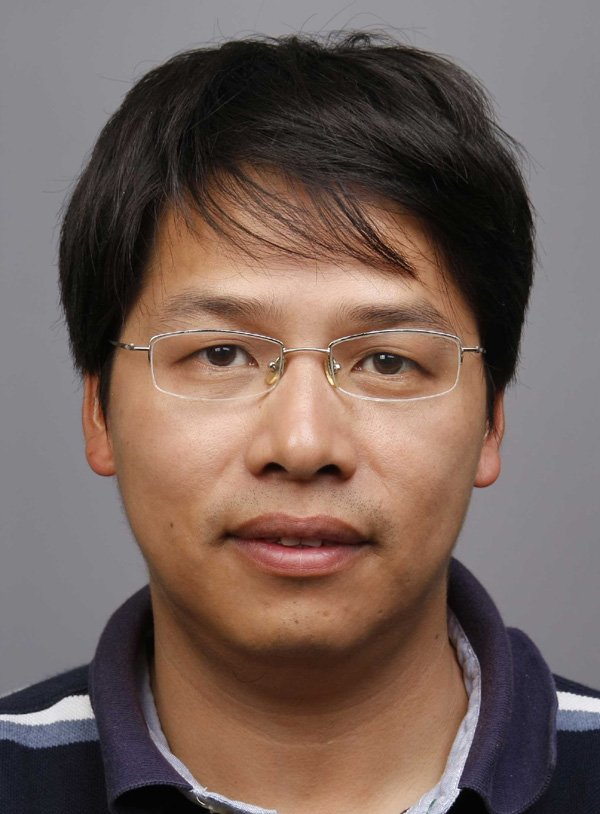}}]{Zhouchen Lin}
received the Ph.D. degree in Applied Mathematics from Peking University, in 2000. He is currently a Professor at Key Laboratory of Machine Perception (MOE), School of Electronics Engineering and Computer Science, Peking University. He is also a Chair Professor at Northeast Normal University and a Guest Professor at Beijing Jiaotong University. Before March 2012, he was a Lead Researcher at Visual Computing Group, Microsoft Research Asia. He was a Guest Professor at Shanghai Jiaotong University and Southeast University, and a Guest Researcher at Institute of Computing Technology, Chinese Academy of Sciences. His research interests include computer vision, image processing, computer graphics, machine learning, pattern recognition, and numerical computation and optimization. He is an Associate Editor of IEEE Trans. Pattern Analysis and Machine Intelligence and International J. Computer Vision, an area chair of CVPR 2014, and a Senior Member of the IEEE.
\end{IEEEbiography}
\begin{IEEEbiography}[{\includegraphics[width=1in,height=1.25in,clip,keepaspectratio]{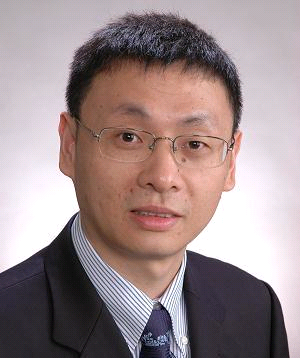}}]{Shuicheng Yan}
is currently an Associate Professor at the Department of Electrical and Computer Engineering at National University of Singapore, and the founding lead of the Learning and Vision Research Group (http://www.lv-nus.org). Dr. Yan's research areas include machine learning, computer vision and multimedia, and he has authored/co-authored hundreds of technical papers over a wide range of research topics, with Google Scholar citation $>$14,000 times and H-index 52. He is ISI Highly-cited Researcher, 2014 and IAPR Fellow 2014. He has been serving as an associate editor of IEEE TKDE, TCSVT and ACM Transactions on Intelligent Systems and Technology (ACM TIST). He received the Best Paper Awards from ACM MM'13 (Best Paper and Best Student Paper), ACM MM’12 (Best Demo), PCM'11, ACM MM’10, ICME’10 and ICIMCS'09, the runner-up prize of ILSVRC'13, the winner prize of ILSVRC’14 detection task, the winner prizes of the classification task in PASCAL VOC 2010-2012, the winner prize of the segmentation task in PASCAL VOC 2012, the honourable mention prize of the detection task in PASCAL VOC'10, 2010 TCSVT Best Associate Editor (BAE) Award, 2010 Young Faculty Research Award, 2011 Singapore Young Scientist Award, and 2012 NUS Young Researcher Award.
\end{IEEEbiography}

%

\end{document}